\documentclass[]{youtu}

\usepackage{mathpazo}
\usepackage{graphicx}
\usepackage{natbib}
\usepackage{amsmath,amssymb}
\usepackage{multirow}
\usepackage{enumitem}
\usepackage{listings}
\usepackage{subcaption}
\usepackage{colortbl}
\usepackage{xspace}
\usepackage{pifont}

\newcommand{\method}{ASAP~}
\newcommand{\passk}[1]{Pass@#1}

\definecolor{rqgray}{gray}{0.94}
\newtcolorbox{rqbox}{
  colback=rqgray,
  colframe=rqgray,
  boxrule=0pt,
  arc=2pt,
  left=6pt,
  right=6pt,
  top=5pt,
  bottom=5pt,
  enhanced,
  breakable
}

\title{Evaluating Agentic Capabilities Before Post-Training: An End-to-End Approach}
\author{Jiarui Qin\textsuperscript{$1 \dagger$}, Junjie Huang\textsuperscript{$2 \dagger$}, Yunjia Xi\textsuperscript{$1$}, Di Yin\textsuperscript{$1$}, Yong Yu\textsuperscript{$2$}, \\ Weinan Zhang\textsuperscript{$2$}, Xing Sun\textsuperscript{$1$}, Weiwen Liu\textsuperscript{$2 \ddagger$}}
\affiliation{$^1$Tencent Youtu Lab \\\textsuperscript{$2$}Shanghai Jiao Tong University}

\usepackage{tabularx}

\usepackage{listings}

\lstnewenvironment{agentlog}
{
    \lstset{
        basicstyle=\ttfamily\footnotesize,
        breaklines=true,
        columns=fullflexible,
        keepspaces=true
    }
}
{}

\begin{document}

\abstract{
Agentic capabilities are increasingly important for modern LLMs, yet remain difficult to evaluate during pretraining. Existing approaches typically rely on static trajectory-derived tasks or likelihood-based proxies, which are lightweight but provide limited visibility into how base models behave in interactive environments.
We introduce \textbf{ASAP (\underline{A}gent \underline{S}caffold \underline{A}daptation for \underline{P}retraining Evaluation)}, a lightweight framework for eliciting and evaluating agentic behavior from pretrained base models on end-to-end tasks. Rather than converting base models into chat-style agents, \method preserves the completion interface and adapts existing agent scaffolds through structured trajectory continuation, short reference prefixes, and minimal recovery mechanisms. Across software-engineering and deep-research benchmarks, \method yields discriminative end-to-end performance across base models and pretraining checkpoints.
Beyond aggregate scores, \method exposes model-generated execution trajectories, enabling analysis of exploration, editing, testing, recovery, and task completion across training stages. These trajectories provide a behavioral view of how agent-relevant capabilities change during pretraining and how execution patterns differ after post-training. Our results suggest that end-to-end agent evaluation can complement conventional pretraining benchmarks by providing a practical diagnostic of interactive model behavior.
}

\maketitle

\vspace{-.1em}

\renewcommand{\thefootnote}{}
\footnotetext{$\dagger$ Equal contribution. jaredqin@tencent.com, huangjunjie2019@sjtu.edu.cn}
\footnotetext{$\ddagger$ Corresponding author.}

% Main sections
\section{Introduction}
\label{sec:intro}

Agentic evaluation is increasingly centered on long-horizon, interactive tasks. Benchmarks such as SWE-bench~\citep{jimenez2024swebench}, DeepSWE~\citep{huang2026deepswe}, and Terminal-Bench~\citep{merrill2026terminalbench} evaluate sustained software-engineering and terminal interaction, while challenging knowledge-oriented benchmarks such as Humanity's Last Exam (HLE)~\citep{phan2025humanity} continue to raise the capability frontier. Yet these evaluations are used almost exclusively \emph{after} post-training.

During pretraining, however, there remains no established way to evaluate how base models behave on end-to-end agentic tasks. Such an evaluation could help track changes across pretraining checkpoints, characterize differences between training stages, and expose behavioral deficiencies that are difficult to observe through conventional static benchmarks alone.

The main challenge is that existing agent benchmarks assume instruction-following models: their scaffolds rely on system prompts, chat templates, prescribed tool formats, and stable multi-turn interaction. Pretrained base models are not explicitly trained for these protocols, and directly inserting them into standard agent scaffolds often produces malformed or unstable trajectories. One alternative is to fine-tune each checkpoint before evaluation, but repeated SFT introduces additional cost and changes the model being evaluated. As a result, existing pretraining-side evaluations largely rely on static or likelihood-based proxies, such as trajectory-derived QA tasks~\citep{qin2025aptbench} or code and trajectory likelihood~\citep{research2026composer2technicalreport,deepseekai2026deepseekv41flash}. These signals are useful for model comparison, but provide limited visibility into how the evaluated model itself interacts with an environment.

This distinction matters because scalar proxy scores cannot show whether a model explores the environment, makes useful edits, tests its hypotheses, revises after failed attempts, or simply fails to make progress. End-to-end interaction therefore provides a complementary view: in addition to final task success, it exposes the execution process through which that outcome is reached. This leads to the central question of our work:

\begin{rqbox}
\centering
\textbf{Can we evaluate pretrained base models on end-to-end agentic tasks without first turning them into instruction-following models?}
\end{rqbox}

We introduce \textbf{ASAP (\underline{A}gent \underline{S}caffold \underline{A}daptation for \underline{P}retraining Evaluation)}, a lightweight framework for adapting existing agent scaffolds to pretrained base models. Our key principle is that \textbf{base models should be evaluated as continuation models rather than incomplete chat models}. ASAP preserves the completion interface and modifies only the model-facing interaction protocol through three simple components: a Markdown+XML trajectory format, a short reference trajectory prefix, and minimal recovery from structural failures such as premature submission or repeated actions.

Across software-engineering and deep-research settings, ASAP yields discriminative end-to-end performance across base models and remains sensitive to differences between checkpoints within the same pretraining run. In our pretraining-stage study, ASAP continues to separate checkpoints even when conventional pretraining benchmarks change only marginally. More importantly, the resulting execution trajectories make it possible to characterize these differences in behavioral terms, including changes in exploration, editing, testing, recovery, and task completion. Comparing base and post-trained models further illustrates how execution patterns differ after post-training.

These results suggest that end-to-end agent evaluation can complement conventional pretraining benchmarks by providing a practical behavioral diagnostic of model behaviors.

Our contributions are:
\begin{itemize}[nosep,leftmargin=*]
    \item \textbf{End-to-end evaluation of pretrained base models.}
    We introduce ASAP, a lightweight adaptation of existing agent scaffolds that enables pretrained base models to execute agentic tasks through their native completion interface, without weight updates.

    \item \textbf{A sensitive probe of agent-relevant behavior during pretraining.}
    ASAP distinguishes both different base models and checkpoints within the same training run, including stages where conventional pretraining benchmarks show only small changes.

    \item \textbf{Trajectory-level diagnosis of behavioral change.}
    ASAP exposes model-generated execution trajectories, enabling fine-grained analysis of how exploration, editing, testing, recovery, and task completion vary across pretraining checkpoints and after post-training.
\end{itemize}
% \begin{figure}[!t]
%     \centering
%     \includegraphics[width=\linewidth]{figures/problem_a2e.pdf}
%     \caption{}
%     \label{fig:problem}
% \end{figure}

\section{Related Work}
\label{sec:related_work}

Existing efforts to evaluate agentic capability before post-training have largely relied on proxy tasks rather than end-to-end agent execution. 
APTBench~\citep{qin2025aptbench} converts successful agent trajectories into static multiple-choice and text-completion problems suitable for base models, while PACE~\citep{song2026pace} predicts agent performance from a set of static benchmark items through cross-model calibration. 
Similarly, model-development pipelines often use coding knowledge, state tracking, or trajectory likelihood as pretraining-side signals. For example, Composer 2~\citep{research2026composer2technicalreport} uses codebase negative log-likelihood for base-model selection, and trajectory BPB/NLL has also been explored as a proxy for downstream agentic capability~\citep{deepseekai2026deepseekv41flash}.

These approaches provide lightweight signals for model selection, but they evaluate fixed examples or reference trajectories rather than the behavior generated by the evaluated model itself. 
ASAP instead adapts existing agent scaffolds so that pretrained base models can execute the original tasks end-to-end through their native completion interface. This enables not only direct evaluation before post-training, but also trajectory-level analysis of how agentic behavior evolves across pretraining checkpoints and is subsequently affected by post-training.
\section{Methodology}
\label{sec:method}
\begin{figure*}[!t]
\centering
\includegraphics[width=0.99\textwidth]{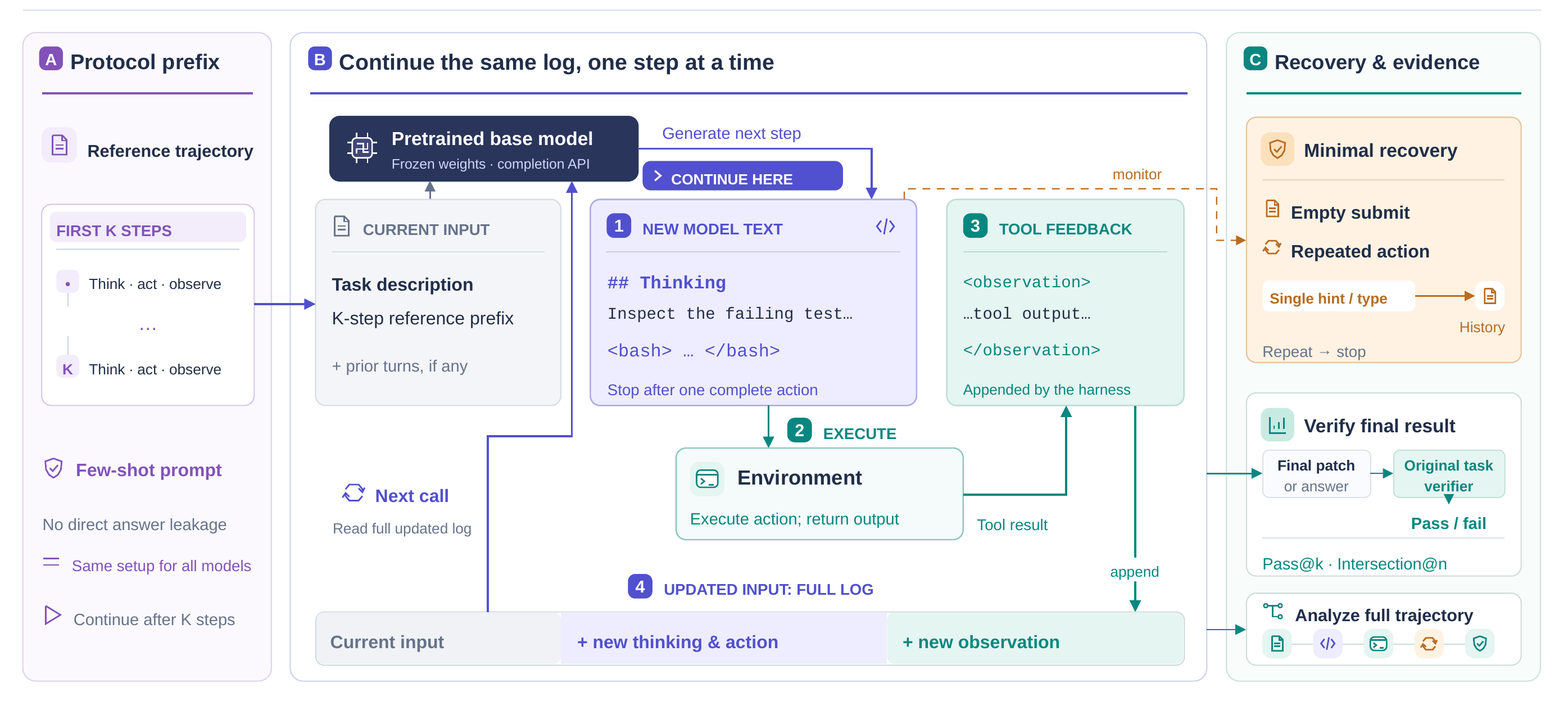}
\caption{\textbf{ASAP enables end-to-end evaluation of pretrained base models.}
\textbf{(A)} A short prefix from a reference trajectory.
\textbf{(B)} The frozen base model continues the trajectory through its native completion interface.
\textbf{(C)} Minimal recovery handles premature empty submissions and repeated actions.}
\label{fig:asap_overview}
\end{figure*}

We evaluate a pretrained base model directly through a simple text completions task, treating agent execution as a continuation problem rather than converting the model into a chat-style interface. 
In particular, we do not apply a chat template or system prompt. Instead, we adapt the underlying agent scaffold with a small set of simple design principles that make base-model generation sufficiently stable for end-to-end agent evaluation. 
\method is built around three components: (1) representing each interaction step in a structured Markdown+XML trajectory format, (2) bootstrapping the model with a short reference prefix that serves as in-context guidance for how to continue the trajectory, and (3) lightweight recovery from structural failures such as repeated actions and empty submit to prevent degenerate loops. 
Figure~\ref{fig:asap_overview} provides an overview of the complete ASAP evaluation pipeline.

\subsection{Markdown and XML Shell-Log Format}
\label{sec:xml}

We formulate agent interaction as a \emph{structured text continuation} task rather than a conversational one.
Each step uses Markdown for free-form thinking and XML tags (e.g., \texttt{<bash>} and \texttt{<observation>}) to delimit actions and environment feedback.
This representation resembles structured text and shell-log patterns commonly seen during pretraining, making it natural for base models to continue.
Among alternative formats including pure JSON, XML, and Markdown, the hybrid Markdown+XML format yields the most stable trajectories in our experiments (Appendix~\ref{apsec:input_format}).
An illustrative format example is provided in Appendix~\ref{apsec:format_example}.

\paragraph{Thought-action generation.}
At each agent step, a single completion call generates both the model's thinking and its subsequent action.
Generation starts from \texttt{\#\# Thinking} and stops after a complete action (e.g., \texttt{</bash>} or a terminal action).
The harness executes the action, appends the resulting output as an \texttt{<observation>} block, and continues the trajectory.
Execution terminates upon \texttt{<submit/>}, the maximum number of steps, or detection of a degenerate repeated-action loop.

\subsection{Reference Trajectory Prefix Bootstrapping}

Unlike instruction-tuned models, pretrained base models are not explicitly trained to initiate a prescribed agent workflow from a standalone task instruction.
We therefore prepend a short segment from a reference agent trajectory, turning the evaluation into continuation of an already established interaction process.
The prefix plays a role similar to few-shot context in conventional base-model evaluation: it helps the model enter the intended interaction pattern and provides lightweight task navigation, while leaving the substantive task solving to the evaluated model.

For each instance, we take the initial steps from a reference trajectory produced by an instruction-tuned model and use them as the bootstrap prefix.
These steps may provide limited task-relevant guidance, such as exposing the repository structure or inspecting potentially relevant files, but we verify that they do not directly reveal the solution or contain the target patch.
The remaining trajectory is then generated autonomously by the evaluated base model.
For each instance, the same prefix is used across all evaluated models, providing a controlled basis for comparison across models and pretraining checkpoints.

Additional details on prefix construction, length ablations, and robustness to reference-source resampling are provided in Appendix~\ref{apsec:prefix_k_ablation}.

\subsection{Lightweight Recovery from Structural Failures}
\label{sec:takeover}

Even with trajectory-format guidance and reference-prefix bootstrapping, base models can occasionally fall into structural failure modes from which they do not recover autonomously. We observe two recurring patterns during harness development: premature submission before making any effective edit, and repeated execution of the same action. We apply a lightweight intervention to each case, providing the model with a single opportunity to recover before terminating the trajectory.

\paragraph{Premature submission recovery.}
Some base models exhibit a \emph{submit bias}, emitting \texttt{<submit/>} before modifying any source file. On the first such attempt, we reject the submission and append a short hint indicating that at least one source file should be edited before submission. The model is then allowed to continue autonomously. If it again attempts to submit without producing an effective edit, we accept the submission as an empty patch and terminate the trajectory.

\paragraph{Repeated-action recovery.}
Another common failure mode is repeated actions, in which the model issues the same action without making progress. 
When such repetition is first detected, we suppress the redundant action and append a short hint asking the model to try a different approach. 
If the model subsequently falls into the same repeated-action pattern again, we terminate the trajectory.

These interventions are deliberately minimal: each failure mode triggers at most one textual hint, and the harness never inserts task-specific actions or edits on behalf of the model. 
If the base model cannot recover after this hint, the trajectory is simply terminated. 
The intervention therefore serves only to prevent trivial structural degeneration from dominating the evaluation, while leaving task solving to the base model itself.

\subsection{Implementation \& Evaluation}
\label{sec:protocol}

\paragraph{Patching Agent Scaffolds.}
We implement our method as lightweight patches to existing agent scaffolds, including SWE-agent~\citep{yang2024sweagent}, OpenHands~\citep{wang2024openhands}, Terminus, and BrowseComp. 
These patches adapt the model-facing interaction loop to support base-model continuation while leaving the underlying task environment and scaffold logic largely unchanged, allowing the same evaluation approach to be applied across different agentic task settings.

\paragraph{Metrics.}
Our primary metric is \passk{k}, computed using the unbiased estimator~\citep{chen2021codex}.
% \begin{equation}
% \passk{k} = \frac{1}{|T|}\sum_{i=1}^{|T|}
% \left[
% 1 - \frac{\binom{n-c_i}{k}}{\binom{n}{k}}
% \right],
% \label{eq:passk}
% \end{equation}
% where $n$ is the number of runs per instance and $c_i$ is the number of successful runs for instance $i$. 
\passk{k} captures whether a task is solvable by the model under repeated sampling. To additionally measure execution stability, we report \textbf{Intersection@$n$}, defined as the number of instances successfully resolved in all $n$ runs. While \passk{k} reflects the model's attainable capability, Intersection@$n$ captures how consistently that capability is realized.

\section{Main Evaluation Results}
\label{sec:experiments}
We first compare six open-source base models on SWE-bench Verified \citep{jimenez2024swebench}, BrowseComp-Plus \citep{chen2025BrowseCompPlus} and Terminal Bench 2.1 \citep{merrill2026terminalbench}. 
We then conduct ablation study of \method and extensive analysis by using \method as an effective pre-training probe.

\subsection{Experimental Setup} \label{sec:exp_setup}
We evaluate six open-source base models spanning the Qwen~\citep{qwen3report,qwen3.5}, GLM~\citep{zeng2025glm}, Gemma~\citep{gemmateam2026gemma4}, and Mimo \citep{mimov25} families. The selection covers same-family size scaling (Qwen3.5-9B vs.\ Qwen3.5-35B), cross-generation comparison (Qwen3-30B vs.\ Qwen3.5-35B), and cross-family comparisons. 
Every model is evaluated on SWEBench Verified, Terminal Bench 2.1 and BrowseComp-Plus using identical \method configuration. 

\subsection{Main Results}
\label{sec:main_results}

\begin{table*}[t]
\caption{
Main results (\%) across agentic benchmarks.
We report \passk{1}, \passk{8}, and the number of instances resolved
in all eight runs (\#Inter).
Full \passk{k} results are provided in
Appendix~\ref{apsec:full_results}.
\textbf{Bold}: best result within each benchmark--scaffold setting.
}
\label{tab:major_main_results}
\vspace{0.5em}
\centering
\small
\setlength{\tabcolsep}{6pt}

\begin{tabular}{lccc@{\hspace{1.5em}}ccc}
\toprule

& \multicolumn{2}{c}{\textbf{SWE-bench Verified}}
& \multicolumn{1}{c}{\textit{SWE-agent}}
& \multicolumn{2}{c}{\textbf{SWE-bench Verified}}
& \multicolumn{1}{c}{\textit{OpenHands}} \\
\cmidrule(lr){2-4}
\cmidrule(l{0.2em}r){5-7}

\textbf{Model}
& \textbf{\passk{1}}
& \textbf{\passk{8}}
& \textbf{\#Inter}
& \textbf{\passk{1}}
& \textbf{\passk{8}}
& \textbf{\#Inter} \\
\midrule

Qwen3-30B-A3B-Base
& 7.38 & 24.20 & 0
& 4.00 & 14.40 & 0 \\

Qwen3.5-9B-Base
& 21.85 & 46.60 & 16
& 32.95 & 55.40 & 51 \\

Qwen3.5-35B-A3B-Base
& 34.23 & 56.20 & 67
& \textbf{39.23} & \textbf{61.80} & 69 \\

Gemma4-26B-A4B-Base
& 7.12 & 23.00 & 0
& 4.67 & 18.20 & 0 \\

GLM-4.5-Base
& 15.97 & 41.40 & 4
& 19.73 & 44.80 & 11 \\

MiMo-V2.5-Base
& \textbf{37.70}
& \textbf{59.40}
& \textbf{82}
& 37.52
& 60.80
& \textbf{70} \\

\midrule
\addlinespace[0.4em]

& \multicolumn{2}{c}{\textbf{Terminal-Bench 2.1}}
& \multicolumn{1}{c}{\textit{Terminus-2}}
& \multicolumn{2}{c}{\textbf{BrowseComp-Plus}}
& \multicolumn{1}{c}{\textit{Native}} \\
\cmidrule(lr){2-4}
\cmidrule(l{0.2em}r){5-7}

\textbf{Model}
& \textbf{\passk{1}}
& \textbf{\passk{8}}
& \textbf{\#Inter}
& \textbf{\passk{1}}
& \textbf{\passk{8}}
& \textbf{\#Inter} \\
\midrule

Qwen3-30B-A3B-Base
& 13.76 & 20.22 & 5
& 17.36 & 32.89 & 29 \\

Qwen3.5-9B-Base
& 19.80 & 30.34 & 10
& 20.53 & 34.94 & 54 \\

Qwen3.5-35B-A3B-Base
& \textbf{26.69}
& \textbf{44.94}
& \textbf{11}
& 26.31
& 43.01
& 97 \\

Gemma4-26B-A4B-Base
& 12.50 & 20.22 & 4
& 14.64 & 30.72 & 20 \\

GLM-4.5-Base
& 16.01 & 33.71 & 6
& 23.34 & 37.47 & 80 \\

MiMo-V2.5-Base
& 25.84
& 41.57
& 10
& \textbf{29.44}
& \textbf{46.51}
& \textbf{125} \\

\bottomrule
\end{tabular}
\end{table*}

Table~\ref{tab:major_main_results} summarizes the performance of six open-weight base models across different agentic benchmarks under \method. Each model is evaluated with eight independent rollouts.

\paragraph{Key observations.}
\textbf{(1) \method provides meaningful discrimination among base models across diverse agentic settings.}
All evaluated models achieve non-trivial end-to-end success, while their performance differs substantially across benchmarks.
On SWE-bench Verified, \passk{1} ranges from 7.12\% to 37.70\% with SWE-agent and from 4.00\% to 39.23\% with OpenHands.
The separation persists beyond software-repository repair: on Terminal-Bench 2.1, \passk{1} ranges from 12.50\% to 26.69\%, while on BrowseComp-Plus it ranges from 14.64\% to 29.44\%.
Qwen3.5-35B-A3B-Base and MiMo-V2.5-Base consistently achieve strong performance across these heterogeneous settings, suggesting that \method captures meaningful differences in base-model agentic capability across both benchmarks and scaffolds.

\textbf{(2) Broad capability tiers are preserved across settings, while exact rankings remain scaffold- and domain-dependent.}
On SWE-bench Verified, the \passk{1} rankings under SWE-agent and OpenHands are strongly correlated (Spearman $\rho \approx 0.89$), and the stronger and weaker models remain broadly separated across both scaffolds.
At the same time, exact rankings and absolute performance can change across scaffolds and benchmarks.
For example, Qwen3.5-9B-Base improves from 21.85\% under SWE-agent to 32.95\% under OpenHands, while Qwen3-30B-A3B-Base decreases; similarly, the relative ordering of Qwen3.5-35B-A3B-Base and MiMo-V2.5-Base varies across Terminal-Bench 2.1, SWE-bench, and BrowseComp-Plus.
These results suggest that agentic capability contains both transferable components and setting-specific behaviors: strong models tend to remain strong overall, but different scaffolds and domains elicit different parts of that capability.
This motivates training for more scaffold-invariant agentic behaviors and broader domain coverage, rather than optimizing for a particular interaction protocol or task distribution.

\textbf{(3) Repeated sampling separates occasional solvability from stable capability.}
The strongest models combine high \passk{1} with large all-run intersections. Under SWE-agent, MiMo-V2.5-Base reaches 37.70\% \passk{1} with 82 instances solved in all eight runs, while Qwen3.5-35B-A3B-Base reaches 34.23\% with an intersection of 67. In contrast, weaker models can benefit substantially from repeated sampling without exhibiting a comparable stable core. Qwen3-30B-A3B-Base, for example, rises from 7.38\% \passk{1} to 24.20\% \passk{8}, yet solves no instance in all eight runs. These results show that multi-run evaluation reveals an important distinction between capabilities that can be elicited occasionally and those that can be executed reliably.

The detailed ablation studies on bootstrap prefix steps, input format design are elaborated in Appendix~\ref{apsec:ablation}.

\section{Extensive Analysis}
In this section, we use ASAP to study how agentic capabilities evolve across pretraining and how they are subsequently affected by post-training. 
We first analyze capability changes across pretraining stages and decompose them into specific behavioral transitions in Section~\ref{sec:probe}, and then examine the changes from base to post-trained models in Section~\ref{sec:pre_to_post}. 
We further find that the model ordering exposed by ASAP is preserved after applying the same lightweight SFT recipe, and the potential evaluation hacking through external network access is scarce for base models. 
These details can be found in Appendix~\ref{apsec:lightsft} and \ref{apsec:hacking}.

\subsection{Monitoring Pretraining Progress and Behavioral Transitions}
\label{sec:probe}

We evaluate checkpoints from different pretraining stages of our self-trained 30B-A3B MoE base model $M$ with ASAP. 
By using \method we could examine how agentic capability evolves over the course of pretraining. 
Specifically, we study whether \method can sensitively distinguish checkpoints at different stages of pretraining, and, beyond aggregate performance gains, identify which agentic behaviors improve as pretraining progresses. 
The latter analysis is enabled by the execution trajectories collected through our end-to-end evaluation: rather than observing only changes in final scores, we can decompose these gains into changes in specific agent behaviors. 
This demonstrates that \method serves not only as a discriminative benchmark for pretraining checkpoints, but also as a diagnostic tool for understanding how agentic capabilities emerge and evolve during pretraining.

\paragraph{Experimental setup.}
We evaluate nine checkpoints spanning five stages with \method on SWE-bench Verified. 
Mid-stage checkpoints are included to measure within-stage progress (stagex-mid means $\frac{2}{3}$ training tokens of stage x). 
Stage1 is general pretraining; stage2 applies learning-rate decay over a higher-quality mixture with QA and instruction-following data but no agentic trajectories; stage3 first introduces trajectory-like data together with 32K long-context training; and stages 4 and 5 continue trajectory training while extending the training context to 128K and 256K, respectively. 
For comparison, we use the aggregate of our conventional pretraining evaluations \footnote{MMLU (-pro,-redux), SimpleQA, INCLUDE, GPQA, GSM8K, MGSM, BBH, MATH500, EvalPlus, Livecodebench, CruxEval, RepoBench, etc}, which cover knowledge \& factual QA, mathematics, reasoning and code.

\begin{table}[t]
\caption{
Training stages of the internal 30B-A3B MoE base model $M$.
Trajectory-like data is first introduced at stage3.
}
\label{tab:training_stages}
\vspace{0.3em}
\centering
\small
\setlength{\tabcolsep}{4pt}

\begin{tabularx}{\linewidth}{
    @{}
    l
    >{\centering\arraybackslash}X
    >{\centering\arraybackslash}X
    >{\centering\arraybackslash}X
    >{\centering\arraybackslash}X
    >{\centering\arraybackslash}X
    @{}
}
\toprule
& \textbf{Stage1}
& \textbf{Stage2}
& \textbf{Stage3}
& \textbf{Stage4}
& \textbf{Stage5} \\
\midrule

Training phase
& General
& LR decay
& \multicolumn{3}{c}{Long-context training} \\

Context length
& 8K
& 8K
& 32K
& 128K
& 256K \\

Trajectory data
& $\times$
& $\times$
& $\checkmark$
& $\checkmark$
& $\checkmark$ \\
\bottomrule
\end{tabularx}
\end{table}

\begin{figure*}[t]
\centering
\includegraphics[width=0.98\textwidth]{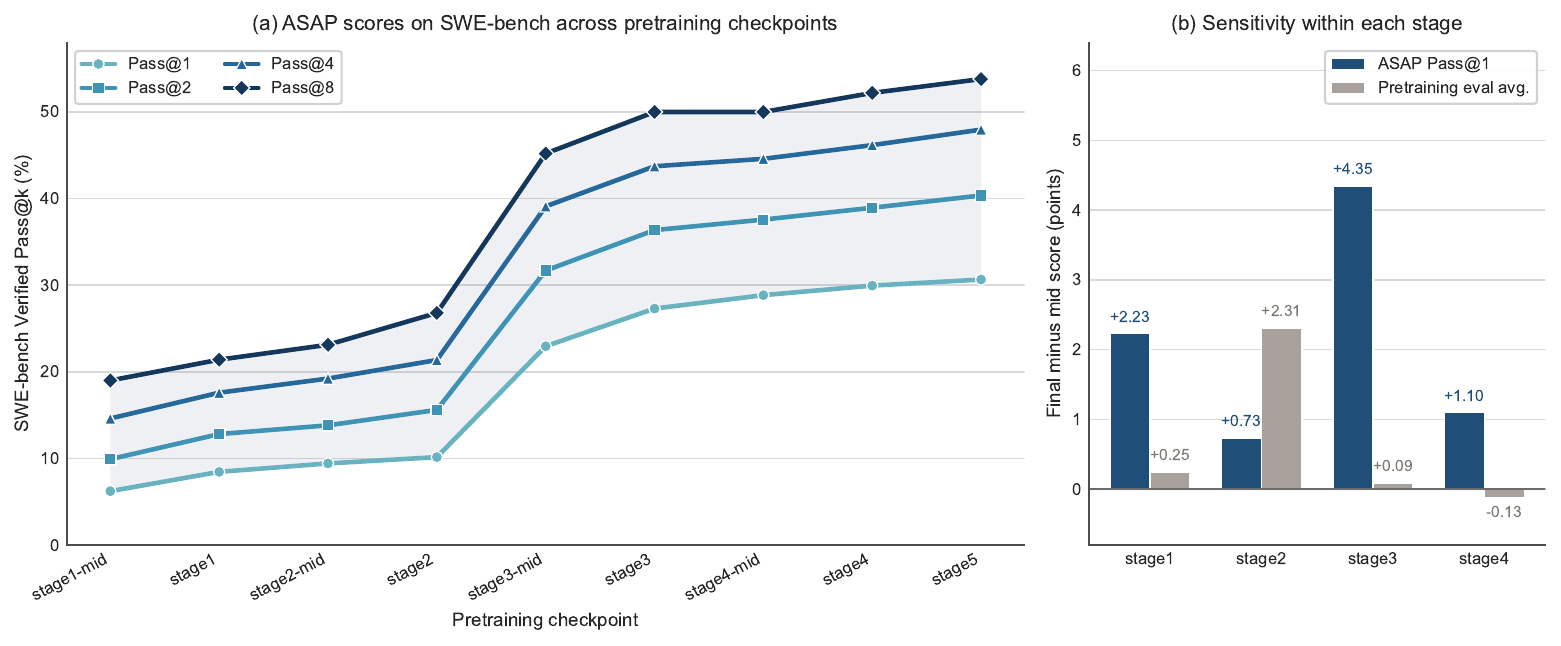}
\caption{Pretraining progress of internal model $M$. \textbf{(a)} \method Pass@k with SWE-agent on SWE-bench Verified across nine checkpoints. \textbf{(b)} \method scores sensitivity within stage1 to stage4, comparing with the conventional pretraining evaluations.}
\label{fig:stages_passk}
\end{figure*}

\subsubsection{Cross-stage performance on \method and general benchmarks}
Figure~\ref{fig:stages_passk}(a) shows how performance on \method evolves across pretraining stages.
The largest transition occurs from stage2 to stage3, where trajectory-like data is introduced together with native 32K long-context training.
While the stage2 model is evaluated beyond its native 8K context using YaRN extrapolation, the stage3 transition still jointly changes both the training data and long-context capability.

Across this transition, \passk{1} increases from 10.15\% at the end of stage2 to 27.30\% at the end of stage3, with stage3-mid already reaching 22.95\%.
Similar improvements are observed across different values of $k$.
Later stages bring more moderate improvements, with \passk{1} reaching 29.95\% at stage4 and 30.65\% at stage5.
Overall, \method sensitively captures the pronounced agent-relevant capability transition around stage3 and the more gradual improvements that follow.

We next examine the within-stage comparisons in Figure~\ref{fig:stages_passk}(b), which provide a more controlled view of how sensitively different evaluations distinguish checkpoints during pretraining. 
\method consistently separates the midpoint and final checkpoint within each stage. 
This distinction is particularly pronounced in stages 1, 3, and 4: \method \passk{1} improves by 2.23, 4.35, and 1.10 percentage points, respectively, whereas the conventional benchmarks scores changes by only +0.25, +0.09, and $-$0.13 points. 
In other words, once training enters the latter half of these stages, the aggregate conventional benchmark score changes only marginally, while \method continues to show measurable improvements.

Stage2 exhibits a different pattern, with the conventional benchmark improving more than \method (+2.31 points versus +0.73pp in \passk{1}). 
Stage2 uses a higher-quality annealing mixture containing QA and instruction-following data together with learning-rate decay, and is therefore specifically optimized to improve conventional knowledge, reasoning, and coding performance without introducing agent trajectory data. 

Taken together, these results show that \method provides strong discrimination both across pretraining stages and among checkpoints within the same stage, and is substantially more sensitive than conventional base-model benchmarks to continued improvements in agentic capability, even in the general pretraining stage (stage1).

\subsubsection{Behavioral decomposition}
\begin{figure*}[b]
\centering
\includegraphics[width=0.99\textwidth]{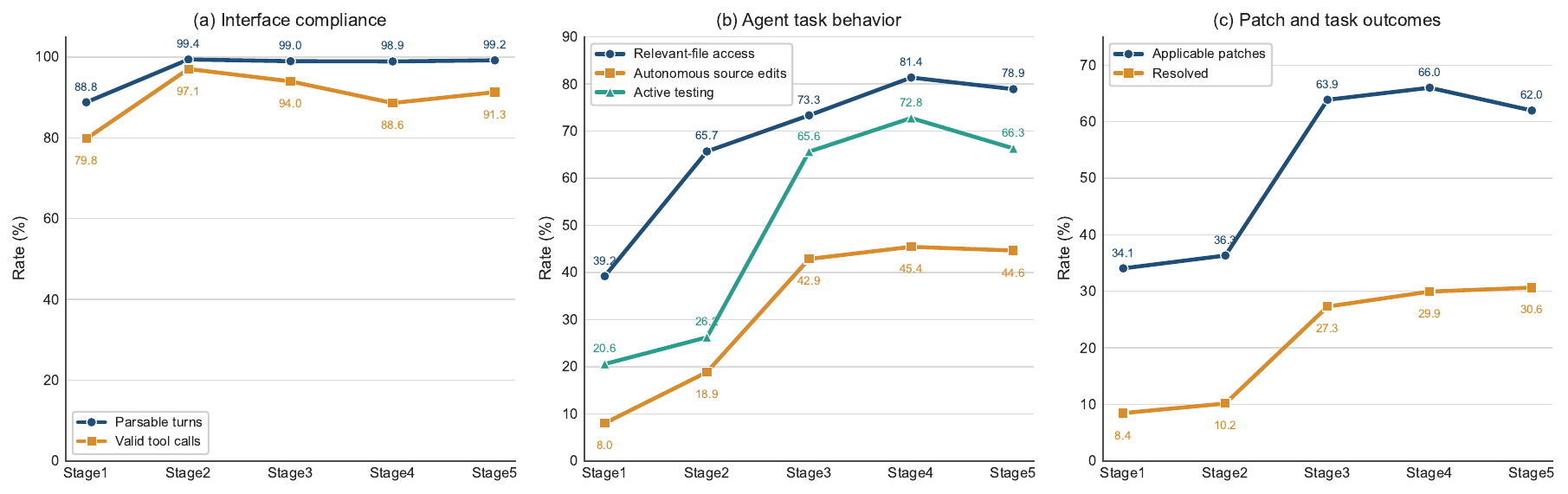}
\caption{Behavioral change across different pretraining stages. Stage2 brings parsability and valid tool calls close to saturation, whereas the main transition in autonomous editing, active testing, applicable patches, and task resolution occurs at stage3.}
\label{fig:stage_behavior}
\end{figure*}

To further investigate what changes at each stage, we compare checkpoints of each stage with the adjacent stages. 
Each pair retains trajectories with the same SWE-bench instance and run index, 4,000 (8 runs) paired trajectories per checkpoint across all 500 tasks; prefix source and prefix length $K$ are also matched. 
We group diagnostics into \emph{interface} behavior (turn parsability, valid tool use), \emph{agentic execution} (relevant-file access, autonomous source edits, active testing), and \emph{outcomes} (applicable and resolving patches). 
The results are reported in Figure~\ref{fig:stage_behavior}.
% reports absolute levels, while adjacent-stage differences use 95\% confidence intervals from 10,000 paired task-level cluster-bootstrap samples.

\paragraph{Stage1$\to$stage2: instruction and interface acquisition.}
Parsable turns increase from 88.81\% to 99.41\%, and valid tool calls from 79.78\% to 97.06\%. Yet only 36.33\% of stage2 trajectories produce an applicable patch and 10.15\% resolve the task.
Together with the introduction of QA/instruction-following data and the absence of agentic trajectories, this separation indicates that stage2 primarily improves interface reliability rather than agentic abilities. 
Relevant-file access also improves, but autonomous edits and active testing improves in a relatively slow manner, resulting in modest resolved rate improvements.

\paragraph{Stage2$\to$stage3: agentic task execution.}
After trajectory data and 32K training are introduced, parsability and valid tool calls remain high, but autonomous edits rise to 42.86\%, active testing to 65.64\%, applicable patches to 63.88\%, and resolution to 27.30\%. 
We also find that repetitive command loops and abnormal termination also become substantially less common in stage3. 
The score increase is accompanied primarily by improvements in execution behaviors rather than interface compliance.

% \paragraph{Stages 3--5: refinement and saturation.}
% From stage3 to stage4, relevant-file access reaches 81.37\%, active testing 72.78\%, and resolution 29.13\%, despite valid tool calls declining to 88.63\%. With matched evaluation limits, these opposing trends suggest improved repository-context integration and continued trajectory refinement rather than better interface compliance. Stage5 shows no statistically significant resolution gain: relevant-file access, active testing, and applicable patches fall to 78.91\%, 66.33\%, and 61.98\%, respectively, while conditional success among applicable patches improves. This pattern suggests a coverage--quality trade-off after the main stage3 transition.

Stages 4 and 5 exhibit modest changes across the individual capability dimensions, which is consistent with the smaller improvement in resolved rate observed at these stages. 
More broadly, these results highlight an important advantage of \method: it allows us to decompose aggregate performance changes into more fine-grained behavioral changes.
Such trajectory-level analysis provides a more informative view of how capabilities evolve during pretraining, and can in turn help diagnose deficiencies and guide the design of pretraining data. 
This type of behavioral attribution is fundamentally unavailable to conventional accuracy-based pretraining benchmarks, which observe only the correctness of the final prediction.

\subsection{Behavioral Differences Before and After Post-Training}
\label{sec:pre_to_post}

We compare base and post-trained models from two families, Qwen3.5-35B-A3B and MiMo-V2.5, on SWE-bench Verified.
Base models are evaluated with \method, while post-trained models use the original SWE-agent scaffold.
Because the protocols differ by design, this comparison is descriptive rather than a scaffold-controlled estimate of the effect of post-training.
Table~\ref{tab:pre_to_post} summarizes the behavioral differences.

\begin{table*}[h]
\caption{
Behavioral differences between base and post-trained models under their respective evaluation protocols.
$\Delta$ is in percentage points.
}
\label{tab:pre_to_post}
\vspace{0.5em}
\centering
\small
\begin{tabular}{lrrr|rrr}
\toprule
& \multicolumn{3}{c|}{\textbf{Qwen3.5-35B-A3B}} &
  \multicolumn{3}{c}{\textbf{MiMo-V2.5}} \\
\textbf{Metric} & \textbf{Base} & \textbf{Post} & \textbf{$\Delta$}
& \textbf{Base} & \textbf{Post} & \textbf{$\Delta$} \\
\midrule
\passk{1} / success rate                & 34.23 & 49.40 & +15.17 & 37.70 & 74.00 & +36.30 \\
Valid source edit                       & 69.99 & 79.41 &  +9.42 & 62.12 & 96.34 & +34.22 \\
Active testing                          & 85.90 & 96.80 & +10.90 & 76.80 & 99.77 & +22.97 \\
Valid re-edit after a failed test       & 31.80 & 63.78 & +31.98 & 28.28 & 57.27 & +28.99 \\
Explicit submission                     & 20.57 & 52.17 & +31.60 & 26.81 & 89.02 & +62.21 \\
Empty patch                             & 22.63 & 13.04 &  $-$9.59 & 34.67 &  1.60 & $-$33.07 \\
Exact match to gold modified-file set   & 44.92 & 61.05 & +16.13 & 57.42 & 72.09 & +14.67 \\
\bottomrule
\end{tabular}
\end{table*}

\paragraph{More reliable execution after post-training.}
Across both families, post-trained models exhibit a more reliable edit-test-recover-submit pattern.
They are more likely to revise their implementation after failed tests, produce fewer empty patches, and explicitly submit a final solution.
Their modified-file sets also align more closely with the gold patch scope.
Taken together, these differences suggest that several components of the agent workflow are already present in base models, but are executed more consistently and with better completion reliability in the post-trained models.

\paragraph{Different bottlenecks in base models.}
The two base models nevertheless exhibit different failure patterns.
Qwen-Base already performs source editing and active testing in a large fraction of trajectories, suggesting that the basic execution workflow can often be elicited before post-training; its main weaknesses lie in recovery, refinement, and reliable completion.
MiMo-Base, in contrast, more often remains in the exploration stage and fails to transition into effective editing.
These patterns illustrate how \method can reveal not only differences in overall task success, but also where the execution pipeline breaks down for different pretrained models.
\section{Conclusion}

We introduced \textsc{ASAP}, a lightweight framework for directly evaluating pretrained base models on end-to-end agentic tasks without weight updates or conversion to a chat-style interface. 
By adapting existing agent scaffolds through structured continuation, short trajectory prefixes, and minimal recovery mechanisms, \textsc{ASAP} enables stable and discriminative evaluation across models, checkpoints, and agent environments. 
More importantly, end-to-end execution exposes the models' own interaction trajectories, allowing capability gains during pretraining to be decomposed into concrete behavioral changes in exploration, editing, testing, and recovery, and enabling direct comparison of these behaviors between pre-training and post-training. 
Our results suggest that agent evaluation can serve not only as a downstream benchmark, but also as a practical diagnostic during pretraining, providing richer signals for understanding capability development and guiding future data and training design.

\setcitestyle{numbers,square}
\setcitestyle{square,numbers,comma}
\bibliography{a2e}

\clearpage
\appendix
\newpage
\section{Appendix}
\label{app:impl}

\subsection{Configuration Parameters}\label{app:details}

Table~\ref{tab:config} lists the key configuration parameters used across our evaluations.

\begin{table}[h]
\caption{Key \method configuration parameters.}
\label{tab:config}
\vspace{0.5em}
\centering
\small
\begin{tabular}{ll}
\toprule
\textbf{Parameter} & \textbf{Value} \\
\midrule
Temperature & 0.7 \\
top\_p & 0.95 \\
Completions calls per step & 1 \\
Generated span per step & Thinking + action \\
max\_tokens (per step) & 2,048 \\
max\_steps & 30 \\
Prefix K & 3 \\
max\_observation\_length & 100,000 chars \\
max\_observation\_in\_prefix & 3,000 chars \\
max\_prefix\_tokens & 64,000 \\
\bottomrule
\end{tabular}
\end{table}

\subsection{Illustration of the selected format}
\label{apsec:format_example}
The following snippet is an example of our Markdown+XML format for the base model evaluation.

\begin{agentlog}
## Thinking
I need to find the relevant file and understand the bug...

<bash>
find . -name "*.py" | grep models
</bash>

<observation>
./src/models.py
./tests/test_models.py
</observation>

## Thinking
Let me examine the failing test to understand the expected behavior...

<bash>
cat tests/test_models.py
</bash>
\end{agentlog}

\subsection{Full evaluation results}
\label{apsec:full_results}
In this section, we show all the Pass@k results in Table~\ref{tab:full_results}.

\begin{table*}[h]
\caption{
Main results (\%) across agentic benchmarks.
\#Inter is the number of instances resolved in all 8 runs.
For SWE-bench Verified, we evaluate each model with both SWE-agent and OpenHands;
Terminal-Bench 2.1 uses the Terminus-2 scaffold;
BrowseComp-Plus uses its native scaffold.
\textbf{Bold}: best in panel.
} \label{tab:full_results}
\vspace{0.5em}
\centering
\small
\begin{tabular}{lccccc}
\toprule
\textbf{Model}
& \textbf{\passk{1}}
& \textbf{\passk{2}}
& \textbf{\passk{4}}
& \textbf{\passk{8}}
& \textbf{\#Inter} \\
\midrule

% ============================================================
% SWE-bench Verified: SWE-agent
% ============================================================
\multicolumn{6}{l}{
    \textbf{SWE-bench Verified}
    \quad \textit{Scaffold: SWE-agent}
} \\
Qwen3-30B-A3B-Base
    & 7.38 & 12.32 & 18.37 & 24.20 & 0 \\
Qwen3.5-9B-Base
    & 21.85 & 30.61 & 38.81 & 46.60 & 16 \\
Qwen3.5-35B-A3B-Base
    & 34.23 & 42.82 & 50.15 & 56.20 & 67 \\
Gemma4-26B-A4B-Base
    & 7.12 & 11.84 & 17.60 & 23.00 & 0 \\
GLM-4.5-Base
    & 15.97 & 24.46 & 33.47 & 41.40 & 4 \\
Mimo-V2.5-Base
    & \textbf{37.70} & \textbf{46.51}
    & \textbf{53.80} & \textbf{59.40} & \textbf{82} \\

\midrule

% ============================================================
% SWE-bench Verified: OpenHands
% ============================================================
\multicolumn{6}{l}{
    \textbf{SWE-bench Verified}
    \quad \textit{Scaffold: OpenHands}
} \\
Qwen3-30B-A3B-Base
    & 4.00 & 6.64 & 10.11 & 14.40 & 0 \\
Qwen3.5-9B-Base
    & 32.95 & 42.04 & 49.32 & 55.40 & 51 \\
Qwen3.5-35B-A3B-Base
    & \textbf{39.23} & \textbf{48.52}
    & \textbf{55.79} & \textbf{61.80} & 69 \\
Gemma4-26B-A4B-Base
    & 4.67 & 8.05 & 12.70 & 18.20 & 0 \\
GLM-4.5-Base
    & 19.73 & 28.58 & 37.07 & 44.80 & 11 \\
Mimo-V2.5-Base
    & 37.52 & 46.99 & 54.53 & 60.80 & \textbf{70} \\

\midrule

% ============================================================
% Terminal-Bench 2.1: Terminus-2
% ============================================================
\multicolumn{6}{l}{
    \textbf{Terminal-Bench 2.1}
    \quad \textit{Scaffold: Terminus-2}
} \\
Qwen3-30B-A3B-Base
    & 13.76 & 16.65 & 18.68 & 20.22 & 5 \\
Qwen3.5-9B-Base
    & 19.80 & 23.52 & 27.35 & 30.34 & 10 \\
Qwen3.5-35B-A3B-Base
    & \textbf{26.69} & \textbf{33.19} & \textbf{39.78} & \textbf{44.94} & \textbf{11} \\
Gemma4-26B-A4B-Base
    & 12.50 & 15.33 & 17.40 & 20.22 & 4 \\
GLM-4.5-Base
    & 16.01 & 20.22 & 25.75 & 33.71 & 6 \\
Mimo-V2.5-Base
    & 25.84 & 31.58 & 36.34 & 41.57 & 10 \\

\midrule

% ============================================================
% BrowseComp-Plus: Native scaffold
% ============================================================
\multicolumn{6}{l}{
    \textbf{BrowseComp-Plus}
    \quad \textit{Scaffold: Native}
} \\
Qwen3-30B-A3B-Base
    & 17.36 & 22.89 & 27.79 & 32.89 & 29 \\
Qwen3.5-9B-Base
    & 20.53 & 26.26 & 31.00 & 34.94 & 54 \\
Qwen3.5-35B-A3B-Base
    & 26.31 & 32.22 & 37.60 & 43.01 & 97 \\
Gemma4-26B-A4B-Base
    & 14.64 & 20.19 & 25.50 & 30.72 & 20 \\
GLM-4.5-Base
    & 23.34 & 28.68 & 33.11 & 37.47 & 80 \\
Mimo-V2.5-Base
    & \textbf{29.44} & \textbf{35.32}
    & \textbf{40.69} & \textbf{46.51} & \textbf{125} \\

\bottomrule
\end{tabular}
\end{table*}

\subsection{Ablation Studies}
\label{apsec:ablation}
This section provides the details and ablation studies of the major design of \method: the format choice and the bootstrap prefix.

\subsubsection{Input Format}
\label{apsec:input_format}

The input representation can substantially affect the ability of pretrained base models to interact with an agent scaffold.
Unlike instruction-tuned models, base models are not explicitly trained to follow a particular tool-use protocol, and therefore the choice of serialization format may introduce a non-trivial source of variance.
We study four commonly used representations: pure JSON, pure Markdown, pure XML, and our default Markdown+XML format.
The latter uses Markdown for free-form reasoning and natural-language context, while reserving XML tags for structured actions and observations.

We conduct the ablation on a randomly sampled 100-instance subset of SWE-bench Verified using the SWE-agent scaffold.
For each format, we sample three independent trajectories per instance and compute \passk{1} and \passk{3} using the unbiased Pass@$k$ estimator.

\begin{table}[h]
\caption{
Input-format ablation on a randomly sampled 100-instance subset of
SWE-bench Verified using the SWE-agent scaffold.
We sample three trajectories per instance and report \passk{1} and
\passk{3} using the unbiased Pass@$k$ estimator.
}
\label{tab:format_ablation}
\vspace{0.5em}
\centering
\small
\setlength{\tabcolsep}{8pt}

\begin{tabular}{llcc}
\toprule
\textbf{Model} & \textbf{Format}
& \textbf{\passk{1}}
& \textbf{\passk{3}} \\
\midrule

\multirow{4}{*}{Qwen3.5-35B-A3B}
& JSON         & 4.33  & 13.00  \\
& Markdown     & 21.67 & 35.00 \\
& XML          & 27.00 & 38.00 \\
& Markdown+XML & \textbf{33.00} & \textbf{50.00} \\

\midrule

\multirow{4}{*}{MiMo-V2.5-Base}
& JSON         & 14.67 & 30.00 \\
& Markdown     & 30.67 & 47.00 \\
& XML          & 38.00 & 52.00 \\
& Markdown+XML & \textbf{42.00} & \textbf{59.00} \\

\bottomrule
\end{tabular}
\end{table}

As shown in Table~\ref{tab:format_ablation}, the representation format has a substantial effect on agent performance.
Pure JSON performs particularly poorly for both models, suggesting that requiring the entire interaction to conform to a rigid structured representation is difficult for pretrained base models.
Markdown performs considerably better, but remains consistently below formats that explicitly delimit structured interaction fields with XML tags.

Pure XML further improves performance, indicating that explicit structural boundaries are useful for separating actions, observations, and other components of the agent trajectory.
However, combining Markdown with XML performs best for both models.
Compared with pure XML, Markdown+XML improves \passk{1} from 27.00\% to 33.00\% for Qwen3.5-35B-A3B and from 38.00\% to 42.00\% for MiMo-V2.5-Base.
The same trend is also reflected in \passk{3}.

These results suggest that Markdown and XML provide complementary benefits.
Markdown offers a natural continuation space for free-form reasoning and textual context, while XML provides explicit local structure for tool actions and observations.
We therefore adopt Markdown+XML as the default representation throughout our experiments.

\subsubsection{Prefix Length and Source Robustness}
\label{apsec:prefix_k_ablation}

\paragraph{Prefix construction.}
For each instance, we construct the reference prefix from trajectories generated by instruction-tuned models.
Our trajectory pool contains trajectories from Kimi K2.5 \citep{kimiteam2026kimik25visualagentic}, GLM-5.2 \citep{glm5team2026glm5vibecodingagentic}, and MiniMax M2.7 \citep{chen2026minimax}.
We randomly select one trajectory and use its initial interaction steps as the prefix.
The prefix is intended to provide \emph{protocol-level} few-shot guidance and demonstrates the expected turn structure, tool-use convention, and Markdown+XML interaction format.
For the default $K=3$ setting, the prefix remains in the initial interaction phase and does not contain a golden patch modification or reveal the ground-truth directly.
The evaluated base model therefore performs the subsequent task-solving trajectory autonomously from step $K+1$ onward.

\paragraph{Prefix-length ablation.}
We evaluate $K\in\{0,1,3\}$ on 100 randomly sampled instances from SWE-bench Verified, using three independent runs to report \passk{1} and \passk{3}.
As shown in Table~\ref{tab:prefix_k_ablation}, increasing the number of prefix steps consistently improves performance for both models.
This is expected for pretrained base models, as additional few-shot context provides a clearer demonstration of the interaction protocol and makes the intended agent workflow easier to continue.

The same trend is observed for both Qwen3.5-35B-A3B and MiMo-V2.5-Base across different values of $K$.
Moreover, their relative ordering remains unchanged: MiMo-V2.5-Base consistently outperforms Qwen3.5-35B-A3B under all prefix lengths and for both \passk{1} and \passk{3}.
This suggests that the prefix mainly improves capability elicitation while preserving the relative comparison between base models.

We use $K=3$ as the default setting.
In preliminary experiments with longer prefixes ($K>3$), we find that some prefixes begin to expose solution-bearing information, including parts of the gold patch.
We therefore limit the prefix to three interaction steps to provide sufficient few-shot guidance while avoiding direct solution leakage.

\begin{table}[h]
\centering
\small
\setlength{\tabcolsep}{8pt}
\caption{
Prefix-length ablation on 100 randomly sampled SWE-bench Verified instances.
Results (\%) are computed from three independent runs.
Increasing the prefix length consistently improves performance while preserving the relative model ordering.
}
\label{tab:prefix_k_ablation}
\begin{tabular}{lccc}
\hline
Model & Prefix length $K$ & \passk{1} & \passk{3} \\
\hline
\multirow{3}{*}{Qwen3.5-35B-A3B}
    & 0 & 24.00 & 41.00 \\
    & 1 & 27.67 & 43.00 \\
    & \textbf{3 (selected)} & 33.00 & 50.00 \\
\hline
\multirow{3}{*}{MiMo-V2.5-Base}
    & 0 & 28.00 & 46.00 \\
    & 1 & 32.00 & 53.00 \\
    & \textbf{3 (selected)} & 42.00 & 59.00 \\
\hline
\end{tabular}
\end{table}

\paragraph{Robustness to prefix source.}
We next examine whether the evaluation conclusions depend on the particular set of reference prefixes used.
We compare two independently constructed prefix sets while fixing the prefix length to $K=3$.
Prefix A is constructed using our default strategy, whereas Prefix B is newly generated by GLM-5.3-Flash for the same set of evaluation instances.

We evaluate Qwen3.5-35B-A3B and MiMo-V2.5-Base on the same 100-instance subset of SWE-bench Verified used in the prefix-length ablation.
For each model--prefix combination, we sample three independent trajectories per instance under the same SWE-agent scaffold, decoding configuration, and interaction budget.
Table~\ref{tab:prefix_source_robustness} reports \passk{1}, \passk{3}, and the score difference between Prefix B and Prefix A.

\begin{table}[t]
\caption{
Robustness to independently constructed reference prefixes on the same
100-instance subset of SWE-bench Verified.
Prefix A denotes our default prefix strategy, while Prefix B is independently
generated by GLM-5.3-Flash.
$\Delta$ denotes Prefix B minus Prefix A.
}
\label{tab:prefix_source_robustness}
\vspace{0.5em}
\centering
\small
\setlength{\tabcolsep}{5pt}

\begin{tabular}{llrrr}
\toprule
\textbf{Model}
& \textbf{Metric}
& \textbf{Prefix A}
& \textbf{Prefix B}
& \textbf{$\Delta$} \\
\midrule

Qwen3.5-35B-A3B
& \passk{1}
& 33.00
& 34.00
& $+1.00$ \\

Qwen3.5-35B-A3B
& \passk{3}
& 50.00
& 48.00
& $-2.00$ \\

\addlinespace[0.2em]

MiMo-V2.5-Base
& \passk{1}
& 42.00
& 44.33
& $+2.33$ \\

MiMo-V2.5-Base
& \passk{3}
& 59.00
& 56.00
& $-3.00$ \\

\bottomrule
\end{tabular}
\end{table}

Changing the reference prefixes produces only modest changes in absolute
performance.
Across the two models and two metrics, the maximum absolute difference between
Prefix A and Prefix B is 3 percentage points.
Moreover, the relative model comparison remains unchanged:
MiMo-V2.5-Base outperforms Qwen3.5-35B-A3B under both prefix sets.

These results indicate that the model separation observed by \method is not
specific to the particular reference prefixes used in our default
configuration.
Instead, independently generated prefixes that establish the same interaction
protocol yield similar absolute performance and preserve the same relative
model ordering.

% To verify that the evaluation is not sensitive to a particular reference model or trajectory, we construct an alternative prefix set by independently re-sampling the reference source for each instance.
% Specifically, for every instance, we randomly select one available trajectory from Kimi K2.5, GLM-5.2, or MiniMax M2.7, and use its first $K=3$ interaction steps as the prefix.
% We then repeat the evaluation with this independently sampled prefix set.
% The resulting performance is similar to that obtained with the original prefix assignment, and the qualitative conclusions remain unchanged.
% This suggests that the measured performance is not driven by idiosyncratic information from a particular reference model or trajectory, but is robust to which valid protocol demonstration is used.

\subsection{Consistency with Lightweight SFT}
\label{apsec:lightsft}
To examine whether the model ordering exposed by direct base-model evaluation
remains informative after instruction following is introduced, we apply the same
lightweight 70K-example SFT recipe to four base models. The goal of this experiment
is not to maximize downstream performance, but to provide a controlled comparison
under a shared post-training recipe. We evaluate each post-SFT model with SWE-agent
for three runs on all 500 SWE-bench Verified instances.

\begin{table}[h]
\caption{Direct base-model evaluation with \method and end-to-end SWE-agent evaluation after the same 70K-example lightweight SFT. Base results use eight runs and SFT results use three runs. All values are percentages. \textbf{Bold}: best in column.}
\label{tab:sft_consistency}
\vspace{0.5em}
\centering
\small
\begin{tabular}{lcc|cc}
\toprule
& \multicolumn{2}{c|}{\textbf{Base / \method}} & \multicolumn{2}{c}{\textbf{70K SFT}} \\
\textbf{Model} & \textbf{\passk{1}} & \textbf{\passk{8}} & \textbf{\passk{1}} & \textbf{\passk{3}} \\
\midrule
Qwen3-30B-A3B       & 7.38  & 24.20 & 19.60 & 31.80 \\
Qwen3.5-9B          & 21.85 & 46.60 & 24.60 & 37.60 \\
Qwen3.5-35B-A3B     & \textbf{34.23} & \textbf{56.20} & \textbf{37.47} & \textbf{53.20} \\
Gemma4-26B-A4B      & 7.12  & 23.00 & 18.70 & 31.60 \\
\bottomrule
\end{tabular}
\end{table}

The model ordering is identical across base Pass@1, base Pass@8, post-SFT Pass@1,
and post-SFT Pass@3, yielding Spearman $\rho=1.0$ for all four pairwise comparisons.
The corresponding Pearson correlations are also positive.
Given the small number of models and the potentially large effect of post-training,
we treat this result only as an aggregate consistency check rather than evidence that
\method fully predicts post-trained performance.
The primary use of \method remains direct evaluation and behavioral analysis of
pretrained base models, as studied in Section~\ref{sec:probe} and Section~\ref{sec:pre_to_post}.

\subsection{External Network Access Audit}
\label{apsec:hacking}
In this section, we investigate the evaluation hacking by external network access of base models with \method.
If an execution environment permits outbound network requests, a model could in principle search for the original issue, pull request, or a known patch instead of deriving the fix from the local repository. 
We therefore audit this observable shortcut by scanning the executable action fields of 4,000 trajectories per model (8 runs $\times$ 500 instances). We count trajectories containing any attempted external network access and separately flag accesses that appear to target a gold solution or external documentation. Table~\ref{tab:hacking_audit} reports the results for the six models with observed network activity.

\begin{table*}[h]
\caption{Trajectory-level audit of attempted external network access. ``Gold'' denotes accesses that appear to seek a reference solution or patch, while ``Docs'' denotes accesses to external documentation.}
\label{tab:hacking_audit}
\vspace{0.5em}
\centering
\small
\begin{tabular}{lrrrrrr}
\toprule
\textbf{Model} & \textbf{Trajectories} & \textbf{Network} & \textbf{Network Rate} & \textbf{Gold} & \textbf{Gold Rate} & \textbf{Docs} \\
\midrule
Qwen3-30B-A3B-Base  & 4,000 & 86 & 2.15\% & 9 & 0.23\% & 23 \\
Gemma4-26B-A4B-Base & 4,000 & 25 & 0.63\% & 5 & 0.13\% & 8 \\
Mimo-V2.5-Base       & 4,000 & 9  & 0.23\% & 1 & 0.03\% & 0 \\
Qwen3.5-35B-A3B-Base      & 4,000 & 9  & 0.23\% & 1 & 0.03\% & 2 \\
Qwen3.5-9B-Base      & 4,000 & 8  & 0.20\% & 1 & 0.03\% & 2 \\
GLM-4.5-Base         & 4,000 & 7  & 0.18\% & 0 & 0.00\% & 6 \\
\midrule
\textbf{Total}       & 24,000 & 144 & 0.60\% & 17 & 0.07\% & 41 \\
\bottomrule
\end{tabular}
\end{table*}

Across the 24,000 audited trajectories, only 0.6\% attempt any network access and 0.07\% are flagged as potentially seeking a gold solution. 
Even for Qwen3-30B-A3B-Base, which has the highest rates, the corresponding values are 2.15\% and 0.23\%. 
Under the deliberately pessimistic assumption that every gold-flagged trajectory succeeds solely because of external lookup, its direct contribution to \passk{1} is therefore at most 0.23 percentage points for any audited model. 
Documentation access is also rare (41 trajectories in total). 
% These results indicate that observable external lookup is not a material driver of the reported scores or model ordering. 
Overall, network activity is rare among base models, indicating that external answer lookup is unlikely to pose a meaningful risk to the validity of our evaluation.

% As with other evaluations of pretrained models, our results inherit potential pretraining-contamination risks from the underlying model benchmark pairs.
% ASAP itself does not attempt to audit or correct such contamination, as it adapts the interaction protocol rather than introducing a new benchmark.

% \subsection{Configuration Parameters}

% Table~\ref{tab:config} lists the key configuration parameters used across all evaluations.

% \begin{table}[h]
% \caption{Key \method configuration parameters.}
% \label{tab:config}
% \vspace{0.5em}
% \centering
% \small
% \begin{tabular}{ll}
% \toprule
% \textbf{Parameter} & \textbf{Value} \\
% \midrule
% Temperature & 0.7 \\
% top\_p & 0.95 \\
% Completions calls per step & 1 \\
% Generated span per step & Thinking + action \\
% max\_tokens (per step) & 2,048 \\
% max\_steps & 30 \\
% Prefix K & 6 \\
% max\_observation\_length & 100,000 chars \\
% max\_observation\_in\_prefix & 3,000 chars \\
% max\_prefix\_tokens & 24,000 \\
% repetition\_window & 3 \\
% repetition\_similarity\_threshold & 0.85 \\
% max\_repetition\_interventions & 3 \\
% Agent workers (parallel) & 8 \\
% Eval workers (parallel) & 6 \\
% \bottomrule
% \end{tabular}
% \end{table}

\subsection{Failure Pattern Analysis}
\label{sec:failure_analysis}

We compare Qwen3.5-35B-A3B-Base and MiMo-V2.5-Base on the same
100 SWE-bench Verified instances, with three independent runs per instance.
Both models use the same SWE-agent scaffold, decoding configuration, and
interaction budget. Table~\ref{tab:failure_patterns} summarizes their main
performance and failure characteristics.

\begin{table}[t]
    \centering
    \small
    \caption{
        Performance and failure patterns on SWEBench Verified using \method.
        ``Non-empty success rate'' denotes the fraction of submitted non-empty
        patches that pass the official grader.
    }
    \label{tab:failure_patterns}
    \begin{tabular}{lcc}
        \toprule
        Metric & Qwen3.5-35B-A3B-Base & MiMo-V2.5-Base \\
        \midrule
        Pass@1
            & 99/300 (33.0\%)
            & \textbf{126/300 (42.0\%)} \\
        Pass@3
            & 50/100 (50.0\%)
            & \textbf{59/100 (59.0\%)} \\
        Empty patch
            & \textbf{73/300 (24.3\%)}
            & 101/300 (33.7\%) \\
        Non-empty unsuccessful/error
            & 128/300 (42.7\%)
            & \textbf{73/300 (24.3\%)} \\
        Non-empty success rate
            & 43.6\%
            & \textbf{63.3\%} \\
        Repetition-based termination
            & \textbf{9/300 (3.0\%)}
            & 24/300 (8.0\%) \\
        Non-empty failures without product-code edits
            & 19
            & \textbf{5} \\
        Non-empty failures with regression
            & 44
            & \textbf{20} \\
        Median failed-patch size (changed lines)
            & 16
            & \textbf{11} \\
        \bottomrule
    \end{tabular}
\end{table}

\paragraph{Distinct execution profiles.}
The two models fail in notably different ways.
Qwen produces non-empty patches more often, but these patches are substantially
less precise: only 43.6\% of its non-empty submissions are correct, compared
with 63.3\% for MiMo.
Its failed patches are also larger and more frequently introduce regressions or
fail to modify the relevant product code.
Manual inspection shows a recurring tendency to edit at the wrong abstraction
level, implement only part of the required behavior, or make unnecessarily broad
changes.

MiMo exhibits the opposite pattern.
Once it commits to a source-code modification, its patches are typically smaller
and more likely to satisfy both the target and regression tests.
However, it is also more likely to terminate without producing a patch, often
after prolonged or repetitive exploration.
The main contrast is therefore not simply overall capability, but execution
style: Qwen acts more readily but with lower precision, whereas MiMo is more
selective but sometimes fails to transition from exploration to implementation.

% The solved-instance sets also overlap substantially.
% Among the 64 instances solved by at least one model, 45 are solved by both
% (Jaccard overlap: 70.3\%).
% This suggests that much of the difference lies in how reliably the models turn
% a plausible analysis into a correct repair, rather than in completely disjoint
% sets of solvable problems.

\paragraph{Implications for targeted data synthesis.}
These failure profiles suggest that pretraining data could be tailored to the
dominant weakness of each model.
For Qwen, the main target is \emph{repair precision}: training examples should
favor minimal, correct patches over alternatives that modify the wrong layer,
introduce regressions, or make unnecessarily broad changes.
For MiMo, the main target is \emph{progress under uncertainty}: trajectories
should encourage the model to move from bounded exploration to a concrete
hypothesis, focused edit, validation, and timely submission, while discouraging
repetitive searches with little new information.

More generally, the trajectory-level diagnostics exposed by \method make it
possible to construct model-specific training signals rather than relying on a
single generic trajectory mixture.
For example, candidate continuations from the same intermediate repository state
can be evaluated with the task grader and used to construct preference pairs
that distinguish precise, successful progress from empty, regressive, or
unnecessarily broad continuations.

\end{document}